# On the Spatiotemporal Dynamics of Generalization in Neural Networks

**Zichao Wei**
Saarland University
ziwe00001@stud.uni-saarland.de

## Abstract

Why do neural networks fail to generalize addition from 16-digit to 32-digit numbers, while a child who learns the rule can apply it to arbitrarily long sequences? We argue that this failure is not an engineering problem but a violation of physical postulates. Drawing inspiration from physics, we identify three constraints that any generalizing system must satisfy: (1) Locality—information propagates at finite speed; (2) Symmetry—the laws of computation are invariant across space and time; (3) Stability—the system converges to discrete attractors that resist noise accumulation. From these postulates, we derive—rather than design—the Spatiotemporal Evolution with Attractor Dynamics (SEAD) architecture: a neural cellular automaton where local convolutional rules are iterated until convergence. Experiments on three tasks validate our theory: (1) Parity—demonstrating perfect length generalization via light-cone propagation; (2) Addition—achieving scale-invariant inference from L=16 to L=1,000,000 with 100% accuracy, exhibiting input-adaptive computation; (3) Rule 110—learning a Turing-complete cellular automaton without trajectory divergence. Our results suggest that the gap between statistical learning and logical reasoning can be bridged—not by scaling parameters, but by respecting the physics of computation

## 1 Introduction

Large Language Models (LLMs) have achieved astonishing success across various complex tasks in natural language processing—translation, summarization, code generation, and even passing the bar exam. However, a perplexing phenomenon coexists with this success: on tasks as simple as addition, which human elementary school students can effortlessly master, models completely collapse as soon as the input sequence length exceeds the range of their training data [1], [2], [3]. This is not a gradual degradation in accuracy, but a precipitous drop from near-perfection to random guessing. This catastrophic failure stands in stark contrast to the model's remarkable performance on other tasks.

This phenomenon is known as "Length Generalization Failure". Generalization, by definition, implies the ability to apply rules learned from training data to unseen situations. For the addition task, true generalization means that after learning to add 16-digit numbers, one should be able to correctly calculate sums for sequences of length 100, 1000, or even 1 million. Research in developmental psychology indicates that human children, within years of learning to count, can generalize the successor function to "all possible numbers"—mastering the essence of mathematical induction [4]. This is true "understanding". Current neural networks, however, fail to achieve this: their "success" exists only within the training distribution; once outside, they fail completely.

The severity of this problem is often underestimated. For tasks with exact answers like addition or parity, 95% accuracy does not mean the model has "basically learned it". On the contrary, for a rule that logically admits no exceptions, any accuracy below 100% proves that the model has not learned the rule at all—it has learned something else, a statistical pattern that happens to correlate with the correct answer within the training distribution. A child who understands addition does not occasionally calculate $2 + 2 = 5$. If a model makes such errors, it is not "imperfectly understanding addition", but "completely failing to understand addition".

This contrast suggests a deeper issue: perhaps the root cause of the Transformer's "success" in language modeling and its "failure" in algorithmic reasoning is *one and the same*. Perhaps the very characteristics that make it so successful in language modeling—such as global attention allowing it to see the entire sequence at once, and different layers learning different representations—are precisely what cause its systematic failure in tasks requiring precise logic and long-range causality [5], [6]. The victory of language modeling may well be the curse of reasoning generalization.

In recent years, the community has attempted countless patches to solve this problem: more complex positional encodings [7], [8], [9], attention biases [10], Chain-of-Thought reasoning [11], [12]… These attempts have achieved local success—some methods extending the generalization range from 1x to 2-3x—but they all stop there. This pattern of "partial success, ultimate failure" is reminiscent of a famous lesson in the history of physics: the various modifications attempting to save the theory of Aether, such as Lorentz contraction and the FitzGerald hypothesis, could explain parts of experimental results but failed to provide a unified explanation. Ultimately, physicists realized the problem lay not in the details, but in the *fundamental assumption*—the Aether itself did not exist.

By analogy, perhaps we should ask a more fundamental question: instead of continuing to patch within the Transformer framework, we should step back and consider—what *physical constraints* must a reasoning system satisfy to achieve *lossless generalization*? The answer may lie not in better engineering tricks, but in more correct *postulates*.

## 2 Theoretical Framework

### 2.1 Two Types of Generalization

Before discussing whether neural networks can "generalize", we must first clarify the meaning of "generalization". Two fundamentally different concepts of generalization exist in the literature.

Classical statistical learning theory [13] defines generalization as empirical risk minimization: the goal is to observe the conditional distribution $P(Y \mid X)$ and find a function that minimizes expected risk. The core assumption of this framework is that training and test data come from the same distribution $P$. However, length generalization is inherently an Out-of-Distribution (OOD) problem—if we only train on sequences of length $L \leq 16$, then a test input of length $L = 100$ lies completely outside the support of the training distribution. In this case, the theoretical guarantees of statistical generalization completely fail.

Pearl's causal framework [14] offers another perspective. His "Ladder of Causality" distinguishes three levels: association ($P(Y|X)$), intervention ($P(Y| \operatorname{do}(X))$), and counterfactuals. The key insight is: if we want a model to remain valid under environmental changes (such as changes in length), what we need is not to learn the "surface correlations" of the data, but to generate the "structural equations" of the data—the underlying causal mechanisms. However, Pearl's work mainly focuses on causal inference and does not directly provide a generalization definition applicable to algorithmic tasks.

### 2.2 Ideal Mathematics Unconstrained by Physics

Based on this idea, we provide the following mathematical definition in an ideal state:

**Definition 1 (Causal Generalization)**
Let $E$ be an environmental intervention variable (e.g., sequence length $L$). Assume the data generation process is governed by an environment-invariant ground-truth structural mechanism $f^*$. We say a model $\hat{f}$ achieves causal generalization if and only if it can recover the true causal mechanism under any environment $E$:

$$\forall x \in \mathcal{X}, \forall E : \quad \hat{f}(x, E) = f^*(x) \tag{1}$$

The key here is that $f^*$ is independent of $E$. Statistical generalization seeks low error under a fixed environment $E_{\text{train}}$, while causal generalization seeks structural recovery of the mechanism $f^*$.

This goal naturally leads to the definition of the core computational unit:

**Definition 2 (Global Logical Operator)**
Let $\mathcal{X}$ be the input space and $\mathcal{Y}$ be the output space. A global logical operator $\Phi : \mathcal{X} \to \mathcal{Y}$ is a deterministic map that maps input sequences to output sequences. For deterministic algorithmic tasks, the true logical operator $\Phi^*$ is the target algorithm itself. Here $\Phi^*$ is the mathematical instantiation of the causal mechanism $f^*$ in Definition 1.

In a purely mathematical sense, $\Phi$ is an abstract black-box map. It allows for any form of dependency between input and output, including infinite precision operations or instantaneous dependencies spanning long distances.

The goal of causal generalization can thus be reformulated as: learning a model $\hat{\Phi}$ such that

$$\forall x \in \mathcal{X} : \quad \hat{\Phi}(x) = \Phi^*(x) \tag{2}$$

Similarly, we need to define the locus where this logical operator acts—the space where computation occurs:

**Definition 3 (Computational Spacetime Manifold)**
The locus of computation is defined as a continuous spacetime manifold $\mathcal{M}_E = \mathcal{S}_E \times \mathcal{T}$, where $\mathcal{S}_E$ denotes the spatial dimension (whose metric extends with environment $E$) and $\mathcal{T}$ denotes the time dimension.

Specifically, for neural networks:

- **Spatial Coordinate** $x$: Corresponds to the information unit position (Token Position) of the input sequence.

  $$x \in \{1, 2, ..., L\} \tag{3}$$

  The metric of the spatial dimension $|x_i - x_j|$ represents the separation of information in the input topology.

- **Time Coordinate** $t$: Corresponds to the evolution depth or recurrent step of the computational process.

  $$t \in \{0, 1, ..., T\} \tag{4}$$

  *Depth is Time.* In deep neural networks (like ResNet, Transformer, or RNN), the computation of each layer represents physical time moving forward one step. The evolution of state $t \to t + 1$ must obey causality.

The system state is a field $\boldsymbol{H}(x, t)$ defined on this manifold, representing the local hidden state at time $t$ and location $x$.

In this ideal definition, spacetime is continuous and smooth, allowing the operator $\Phi$ to propagate information across this manifold in any manner (including action at a distance).

### 2.3 Physical Postulates: From Ideal to Realizable

The above definition describes a computational system that is mathematically perfect but physically unrealizable. In this "Creative Mode", the global operator $\Phi$ possesses infinite degrees of freedom and can exploit any long-range correlation of the input sequence—including spurious correlations dependent on specific sequence length $L$. This is precisely the root cause of the Transformer's excellent in-distribution performance but collapse out-of-distribution: excessive degrees of freedom lead to overfitting of environmental variables (like length).

To achieve true causal generalization, we need to introduce constraints to "collapse" this ideal space. What constraints should we choose?

The human brain is currently the only known system capable of achieving perfect length generalization (e.g., applying addition rules to arbitrary lengths after learning). This suggests that the physical mechanism for causal generalization exists in nature.

As [15] profoundly pointed out, biological nervous systems are strictly constrained by physical laws: signal transmission has delays (Time Delays), and neural dynamics manifest as spatiotemporally inseparable traveling waves rather than instantaneous global updates. This gives us a key insight: the spatiotemporal dynamic constraints of the biological brain may not be evolutionary "compromises", but "necessary conditions" for the formation of intelligence. Precisely because it cannot perform "action at a distance", the brain is forced to learn causal mechanisms with locality and translational invariance—mechanisms that naturally possess generalization capabilities.

Therefore, we draw intuition from the spatiotemporal dynamics of the human brain and physics, and abstract them into physical postulates adaptable to artificial neural networks. We analyze these constraints from the perspectives of kinematics and dynamics.

#### 2.3.1 Kinematic Postulates and Constraints

**Postulate 1 — Relativistic Causality**

In physical reality, we never observe true "Action at a Distance". For two spatially separated entities $A$ and $B$ to interact, they must transmit influence through a medium (field or particle) at a finite speed, traversing space and contacting adjacent points sequentially.

Consider a simple physical process: toppling a row of dominoes. The $i$-th domino falls, pushing the $(i + 1)$-th. This is the generative mechanism producing the phenomenon. If the 1st domino falling instantly causes the $L$-th domino to fall—when we speak of "instantly", this is equivalent to the 1st and $L$-th dominoes falling simultaneously. Then, "causation" does not exist; from our side, we might think the 1st caused the $L$-th, but an observer on the other side (or in another reference frame) might think the $L$-th caused the 1st. As relativity reveals, causality becomes undefinable here.

As we know, causality implies strict temporal ordering. "$A$ causes $B$" strictly implies $A$ occurs before $B$. Formally:

$$A \text{ causes } B \quad \Longrightarrow \quad t(A) < t(B) \tag{5}$$

Under action at a distance ($\Delta t = 0, \Delta x \neq 0$), temporal order becomes blurred or even reversible, rendering causality itself physically undefinable.

**Postulate 1 (Relativistic Causality).** To restore the causal structure required for generalization, we introduce the first fundamental postulate: a strict Causal Horizon exists in spacetime.

This can be equivalently expressed as two mutually necessary and sufficient constraints:

- **Kinematic Constraint (Finite Speed)**: The speed of information propagation has an upper bound $c < \infty$.
- **Topological Constraint (Locality)**: The state update of any point can only depend on its neighborhood within its past light cone.

$$\text{Postulate 1:} \quad \frac{\Delta x}{\Delta t} \leq c \Longleftrightarrow \boldsymbol{h}_{t+1}(x) = f(\boldsymbol{h}_t(\mathcal{N}_c(x))) \tag{6}$$

**Postulate 2 — Spacetime Symmetry**

We naturally believe that mathematical and physical laws do not change with time and space. $1 + 1 = 2$ on Earth, and it equals 2 on the Moon; an apple falls today, and it should fall tomorrow. This is the foundation of induction.

If physical laws varied arbitrarily with coordinates $(x, t)$—for example, a carry rule acting at the 10th position changed to a different rule at the 100th position—then the experience gained from finite observations (training set) would have no reason to generalize to unknown spatiotemporal regions (test set).

In physics, this property of "laws invariant to coordinate changes" is called Spacetime Translation Symmetry. As pointed out by [15], according to Noether's Theorem, every continuous symmetry corresponds to a conserved quantity.

Keller et al. argue that this conserved quantity is the basis of "memory" in neural systems. We further point out here that in the context of causal generalization, this symmetry guarantees the conservation of the "causal mechanism" itself. In other words, only when the operator satisfies spacetime symmetry does the model learn a universal Law, rather than a location-specific History.

To ensure the model learns "truth" independent of scale rather than "bias" specific to location, we introduce the second postulate.

**Postulate 2 (Spacetime Symmetry).** We postulate that the evolution operator $f$ must possess Translation Invariance. That is, for any shift $(\Delta x, \Delta t)$ on the spacetime manifold, the system's evolution rule remains unchanged:

$$\text{Postulate 2:} \quad f_{x,t}(\cdot) \equiv f_{x+\Delta x, t+\Delta t}(\cdot) \equiv f_{\text{shared}(\cdot)} \tag{7}$$

The evolution operator $f$ does not depend on absolute coordinates $(x, t)$.

Combining Kinematic Postulate 1 (Relativistic Causality/Locality) and Kinematic Postulate 2 (Spacetime Symmetry), we can now derive the mathematical form that a neural network operator must follow to achieve causal generalization.

From Postulate 1 (Speed Limit), the global state $\boldsymbol{H}_{t+1}$ cannot be obtained by a global map $\Phi(\boldsymbol{H}_t)$, but must be decomposed into functions of local neighborhoods. From Postulate 2 (Symmetry), this local function $f$ must share parameters across all spatiotemporal locations.

Therefore, the global logical operator $\Phi$ in the ideal mathematical definition collapses into a single, weight-shared Local Evolution Operator:

$$\boldsymbol{h}_{t+1}(x) = f(\boldsymbol{h}_t(x-r), ..., \boldsymbol{h}_t(x), ..., \boldsymbol{h}_t(x+r)) \tag{8}$$

Notably, this strictly corresponds mathematically to a **Convolution Kernel** or **Recurrent Rule**.

### 2.3.2 Dynamic Postulates and Constraints

Kinematic postulates establish the geometric structure of computational spacetime: locality dictates legal paths for information transmission, and symmetry dictates physically shared laws across all spacetime. These two postulates jointly define an ideal Channel.

However, possessing a perfect channel does not mean information can be transmitted infinitely far. In the physical world, any sustained evolutionary process faces the challenge of the Second Law of Thermodynamics—Entropy Increase. If the system's evolution operator is linear or merely unitary, initial microscopic noise will be preserved and accumulated over long transmissions, eventually destroying the integrity of causal logic.

To maintain information purity in long-range spacetime, we must introduce dynamic constraints.

**Postulate 3 — Thermodynamic Dissipation and Stability**

Imagine a game of telephone: the first person whispers a sentence, passing it sequentially; by the tenth person, it is often unrecognizable. This is not because someone intentionally changed the information, but because microscopic errors at each step accumulate, eventually overwhelming the original signal.

Analog signals (like videotapes) degrade with each copy because copying introduces noise; digital signals (like DVDs) can be copied losslessly because information is quantized into discrete 0s and 1s—microscopic perturbations are “constrained” back to the nearest legal state rather than accumulating infinitely.

To maintain long-range order in a noisy physical world (or a computer with numerical precision errors), a computational system must possess a Dissipation mechanism capable of expelling noise energy from the system, simplifying states naturally back to ordered orbits.

To achieve theoretically infinite-length causal reasoning, the computational system must possess the ability to counteract entropy increase. We introduce the third postulate.

**Postulate 3 (Thermodynamic Dissipation and Stability).** We postulate that the state evolution operator $f$ must constitute a Dissipative Dynamical System. This means that a low-dimensional Attractor Structure $\mathcal{A}$ must exist in the State Space, such that any state perturbed by microscopic noise converges back to these attractor structures during evolution:

$$\text{Postulate 3: } \lim_{t\to\infty} \operatorname{dist}\big(f^t(\boldsymbol{h}+\varepsilon), \mathcal{A}\big) = 0 \tag{9}$$

The evolution operator $f$ must be dissipative: attractors $\mathcal{A}$ exist in the phase space such that all trajectories asymptotically converge to $\mathcal{A}$.

In neural network implementation, this corresponds to introducing Contractive Nonlinearity (e.g., saturation activation functions or explicit quantization operations).

In the context of discretized algorithmic logic, Postulate 3 imposes the final constraint on the computational medium: although the underlying physical carrier (like GPU floating-point numbers) is continuous, to combat entropy increase, effective computational states must collapse into discrete states.

### 2.4 Redefining

After imposing the three physical postulates, we must revisit and revise the idealized definition in Section 2.1.

**Definition 2′ (Local Evolution Operator).** The global logical operator $\Phi$ collapses into a local evolution operator $f$. Due to Postulate 1 (Locality) and Postulate 2 (Symmetry), $f$ must be a parameter-shared map acting on a finite neighborhood $\mathcal{N}$:

$$\boldsymbol{h}_{t+1}(x) = f(\boldsymbol{h}_t(x-r : x+r)) \tag{10}$$

This is mathematically equivalent to **Convolution**.

**Definition 3′ (Discrete Lattice Space).** The continuous computational manifold $\mathcal{M}$ collapses into a discrete lattice space $\mathcal{L}$. Due to Postulate 3 (Stability), the state space must possess discrete attractor structures, such that effective computation occurs on a discrete symbol set $\mathcal{S}$.

#### 2.4.1 The Isomorphism

Here, we observe a profoundly deep mathematical fact: when we formalize the above physically constrained system, it precisely reconstructs the computational model proposed by Von Neumann in the last century: Cellular Automata [16].

**Corollary 1: Isomorphism to Cellular Automata.** A computational system satisfying locality, translational invariance, and state discreteness is strictly mathematically isomorphic to the 4-tuple definition of Cellular Automata (CA): $\langle \mathcal{L}, S, \mathcal{N}, f \rangle$.

This discovery indicates: Cellular Automata are not purely man-made algorithms, but the inevitable form of any intelligent system attempting to achieve infinite causal generalization under physical constraints.

#### 2.4.2 SEAD: A Perfect Neuro-Symbolic Architecture

This isomorphism points us to the ultimate architecture for solving the generalization problem—a native Neuro-Symbolic Model.

Traditional neuro-symbolic methods are often disjointed (e.g., neural networks parsing images + symbolic solvers reasoning). Our theory derives an organic fusion:

- **The Neural Component (Kinematics / Rule Learning)**: We use Convolutional Neural Networks (CNN) to parametrize the local rule $f$. Through the gradient descent capability of deep learning, we automatically learn complex local causal mechanisms (i.e., "physical laws") from data, solving the pain point that traditional CA rules are hard to design manually.
- **The Symbolic Component (Dynamics / Inference)**: We use the lattice structure and discrete attractors of Cellular Automata (CA) as the inference engine.

By forcing the system to evolve over multiple steps in a discrete symbol space, utilizing attractors to combat noise and lattices to guarantee causality, this system inherits the infinite long-range reasoning capability and perfect generalization of symbolic systems.

Based on this, we propose the SEAD architecture. In the following, we will introduce the specific framework of the SEAD architecture and its performance on relevant tasks.

## 3 SEAD Architecture

In the theoretical section, we established the three physical postulates that neural computation must follow. In this chapter, we formalize these postulates into a general computational architecture—SEAD (Spatiotemporal Evolution with Attractor Dynamics). To simultaneously satisfy the differentiability requirements of deep learning and the stability requirements of symbolic systems, the SEAD architecture adopts a unique Dual-Phase Operation mechanism. We mathematically decouple the training and inference phases, viewing them as a kinematic fitting process and a dynamic evolution process, respectively.

### 3.1 Training Mode: Kinematic Learning on Continuous Manifold

In the training phase, we view SEAD as a pure Convolutional Neural Network (CNN). At this point, the system is only constrained by kinematic postulates (P1 Locality & P2 Symmetry), without introducing dynamic stability constraints for the time being.

Our goal is to learn a parameterized approximation $f_\theta$ of the local physical law $f^*$.

**Continuous Relaxation**: Although the target state is discrete, during training, we operate on the Continuous Probability Simplex. Input $x$ and state $h$ are mapped to a continuous manifold $\mathcal{M} \cong \mathbb{R}^d$ via an embedding layer:

$$\boldsymbol{e}_t = \mathrm{Embed}(h_t, x) \in \mathcal{M} \tag{11}$$

The Neural Kernel calculates the evolution vector field on the manifold. We adopt a **Chaos Training** strategy, directly optimizing single-step transition probabilities in a randomly sampled state space:

$$P_{\theta(h_{t+1} \mid h_t)} = \mathrm{Softmax}(\mathrm{CNN}_\theta(\boldsymbol{e}_t)) \tag{12}$$

Since we optimize probability distributions (Logits) rather than discrete actions, the entire process is fully differentiable. The loss function is defined as the Cross-Entropy between the predicted distribution and the Ground Truth physical law:

$$\mathcal{L} = \mathrm{CrossEntropy}\left(P_\theta, h_{\text{target}}\right) \tag{13}$$

In the training phase, SEAD is a standard, stateless function approximator. It does not concern itself with long-range evolution, focusing only on precisely fitting the "kinematic shape" of the local causal mechanism.

### 3.2 Inference Mode: Dynamic Evolution on Symbolic Lattice

In the inference phase, to achieve infinite long-range causal generalization, we introduce the dynamic postulate (P3 Thermodynamic Stability). The system switches to Cellular Automata mode.

The inference process is an iterative dynamic evolution, which we call the **Relax-and-Project** loop. Each update step includes two sub-steps:

1. **Relax**: Discrete symbols are Lifted to the continuous manifold, and the trained neural kernel $f_\theta$ is applied to calculate update trends:

$$z_{t+1} = \text{CNN}_\theta(\text{Embed}(h_t)) \tag{14}$$

2. **Project**: The continuous update signal is forcibly projected back to the discrete symbol space $\mathcal{S}$. This is mathematically implemented via Hard Decision:

$$h_{t+1} = \text{argmax}(z_{t+1}) \tag{15}$$

At this point, the "Project" operation is non-linear and acts dynamically as a **Denoiser**. Any microscopic numerical error $\varepsilon$ generated during continuous computation, as long as it does not cross the decision boundary, will be essentially zeroed out in the projection step. This ensures that the state trajectory always adheres to **Discrete Attractors**, thereby allowing causal waves to propagate losslessly over infinite distances.

**Convergence Detection**: The stopping condition for inference does not require an external Oracle, but is determined by the system *automatically detecting fixed points*: when $h_{t+1} = h_t$ (i.e., states at all positions no longer change), the system has entered an attractor, and computation naturally terminates. This corresponds to the physical intuition: the wave propagation completes, and the medium returns to rest. This "self-knowing termination" capability is a direct corollary of Postulate 3 (Stability)—attractors not only resist noise but also mark "computation completion".

### 3.3 Summary of the Architecture

This separation of training and inference is key to SEAD as a successful neuro-symbolic model:

Table 1: Training and inference phases of SEAD: mathematical spaces and postulate constraints.

| Phase | Mathematical Space | Postulate Constraints |
|---|---|---|
| Training | Continuous Manifold $\mathcal{M}$ | P1 Locality + P2 Symmetry |
| Inference | Discrete Lattice $\mathcal{S}^L$ | P1 + P2 + P3 Stability |

During training, SEAD leverages the gradient advantages of continuous space to efficiently learn complex non-linear rules. During inference, it leverages the stability advantages of discrete space to achieve noise-resistant long-range symbolic reasoning.

## 4 Experiments

Unlike the pursuit of state-of-the-art results on fixed test sets, this chapter aims to verify whether the SEAD architecture truly learns and follows spatiotemporal causal physical laws through three carefully designed "Physics Probes".

We adopt a strict Out-of-Distribution (OOD) setting in all experiments: models are trained only on short sequences (e.g., $L_{\text{train}} = 16$ or 32) but tested on extremely long sequences ($L_{\text{test}} \to 10^4$ or even higher).

### 4.1 Probe I: High-Sensitivity Propagation (The Parity Task)

The Parity task requires calculating the cumulative XOR of an input sequence $x \in \{0, 1\}^L$. This is a typical *high-sensitivity function*—changing any single input bit flips all outputs from that position to the end of the sequence. Physically, this corresponds to a Soliton propagating losslessly in a medium: any perturbation at position $i$ propagates along the light cone, affecting all states $j > i$.

[17] proved that the loss landscape of Transformers naturally biases towards low-sensitivity functions. This is because high sensitivity requires information to be *transmitted losslessly to arbitrary distances*—precisely the synergistic goal of the three postulates: *Locality* ensures information propagates along the correct causal path (light cone), *Stability* ensures information does not decay during propagation, and *Symmetry* ensures the propagation rule is consistent at every position. Parity is a comprehensive stress test for the three postulates.

As $L \to \infty$, any microscopic analog noise $\varepsilon$, if amplified during propagation (as in RNNs or Transformers), leads to waveform collapse. According to Postulate 3, SEAD's "Relax-and-Project" mechanism should act as a denoiser. We expect the model to propagate the wave losslessly to arbitrary lengths, with the required physical time steps $T$ strictly linearly dependent on $L$ ($T \propto L$).

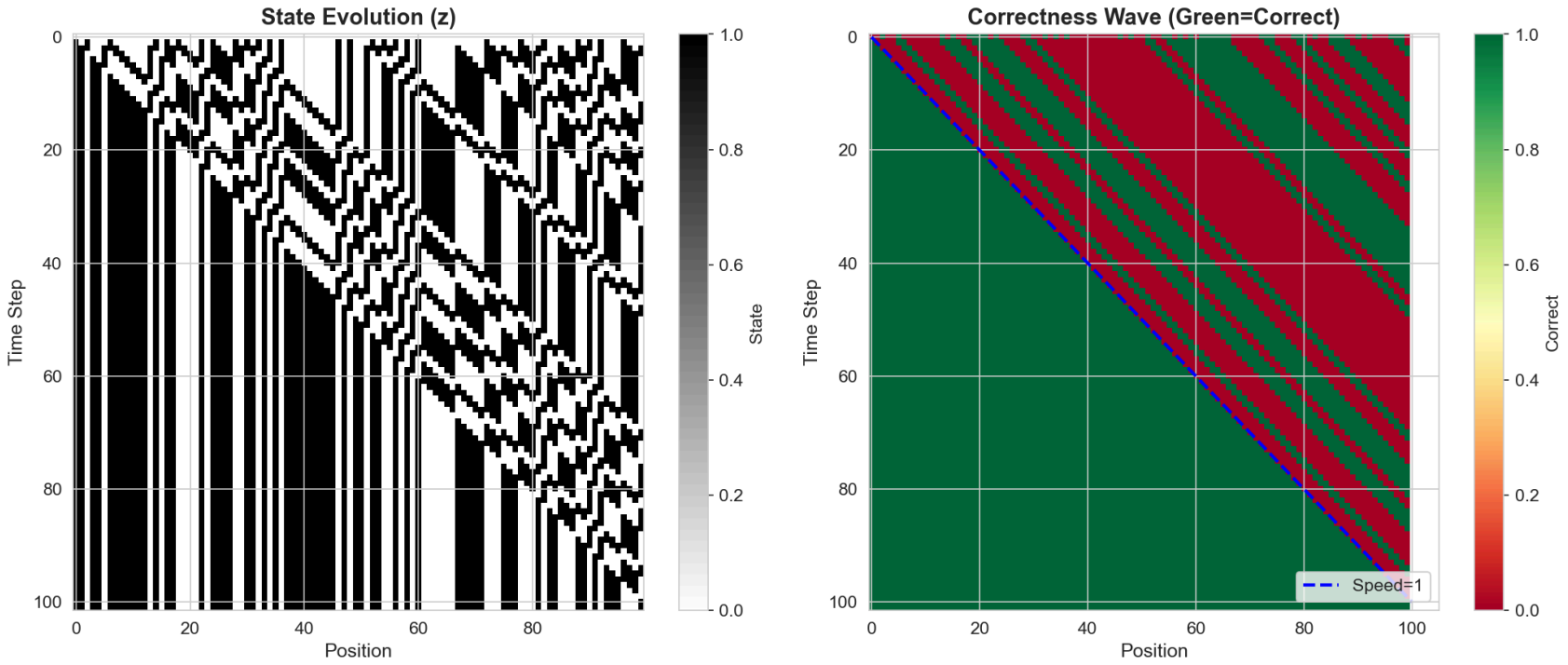

Figure 1: **Spatiotemporal evolution of the Parity task.** Horizontal axis (Position) represents spatial coordinates; vertical axis (Time Step) represents evolution depth, increasing downwards. **Left**: Evolution starting from a random initial state. The cumulative XOR wave propagates from left to right at light speed $c = 1$, gradually ordering the lattice. **Right**: Correctness wave (Green=Correct, Red=Incorrect). The wavefront of correctness strictly follows the causal light cone (blue dashed line, slope=1), empirically validating Postulate 1 (Locality). After approximately $L$ steps, the entire sequence converges to the correct answer.

We trained the model on sequences of length $L = 16$ and tested it on sequences up to $L = 10^6$. The model has only 41 trainable parameters and converges in about 900 steps (~3 seconds) on a home computer (RTX 4080 Super).

The results are shown in the table below. The model achieved perfect length generalization (from 16 to $10^6$, a generalization factor of 62,500x), and the inference steps $T$ were strictly equal to $L + 1$, verifying the theoretical expectation of soliton propagation at light speed $c = 1$.

Table 2: Generalization results for the Parity task. Output is 100% Exact Match.

| Test Length | Samples | Accuracy | Steps |
|---|---|---|---|
| 16 | 1,000 | 100.0% | 17 |
| 64 | 1,000 | 100.0% | 65 |
| 256 | 1,000 | 100.0% | 257 |
| 1,024 | 1,000 | 100.0% | 1,025 |
| 10,000 | 100 | 100.0% | 10,001 |
| 100,000 | 10 | 100.0% | 100,001 |
| 1,000,000 | 1 | 100.0% | 1,000,001 |

### 4.2 Probe II: Algorithmic Adaptivity (The Addition Task)

Binary addition requires computing the sum of two $L$-bit integers. Physically, this is a controlled wave propagation process. The generation, propagation, and annihilation of the Carry Wave depend on the local medium state (input bits $a_i, b_i$).

We compare two extreme cases:

- **Case A (Random)**: Random inputs, short carry chains.
- **Case B (Adversarial)**: Adversarial samples (e.g., $111\cdots1 + 1$), where the carry wave must traverse the entire domain.

A true intelligent physical system should possess a “Least Action” property:

- For Case A, the system should converge rapidly (low entropy, simple task).
- For Case B, the system should automatically prolong inference time (high entropy, complex task).

SEAD exhibits distinct **Adaptive Dynamics**.

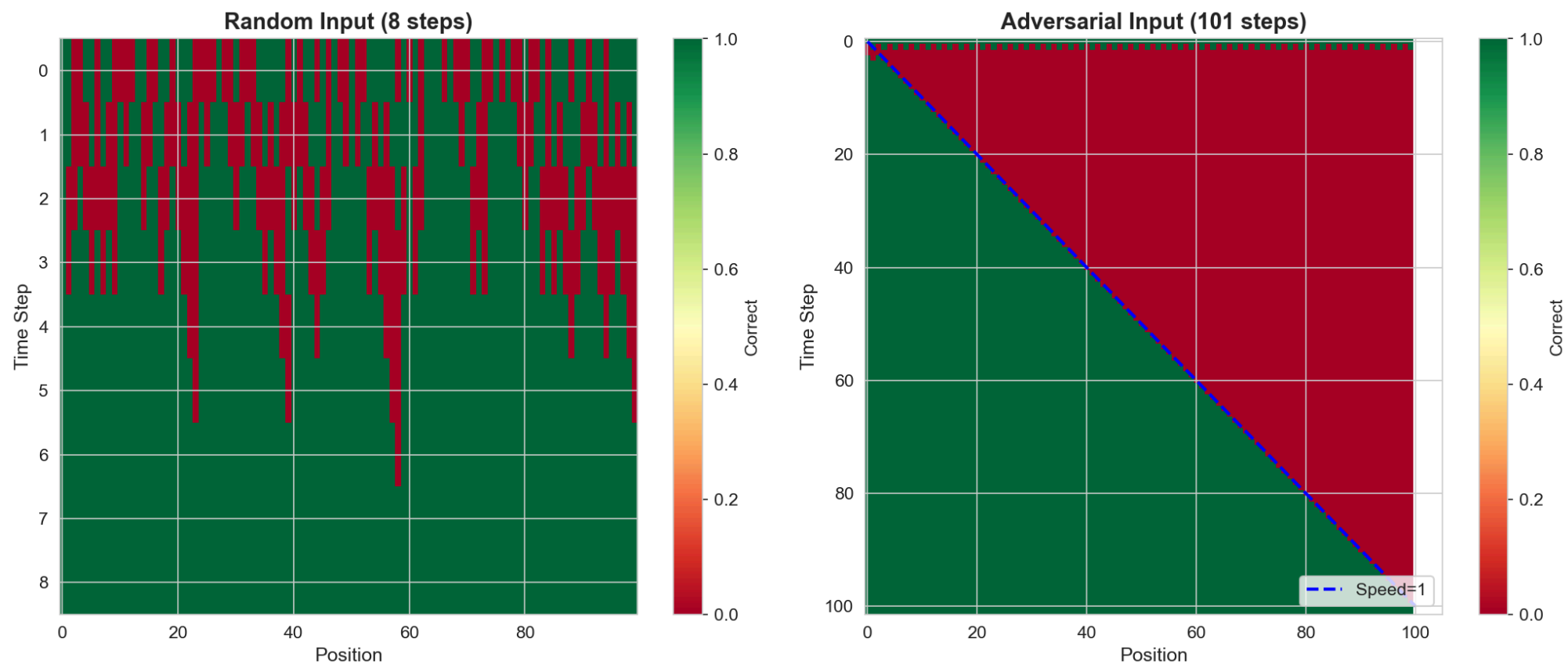

Figure 2: **Spatiotemporal evolution comparison: Random vs Adversarial inputs.** Horizontal axis (Position) represents spatial coordinates; vertical axis (Time Step) represents evolution depth. Green=Correct, Red=Incorrect. **Left**: Random input. Due to short carry chains, full convergence is reached in about 8 steps (all green), showing an “island-like” rapid convergence pattern. **Right**: Adversarial sample $1^L + 1$. The carry wave must traverse the entire domain. The wavefront of correctness strictly follows the causal light cone (blue dashed line, Speed=1), converging after about 101 steps.

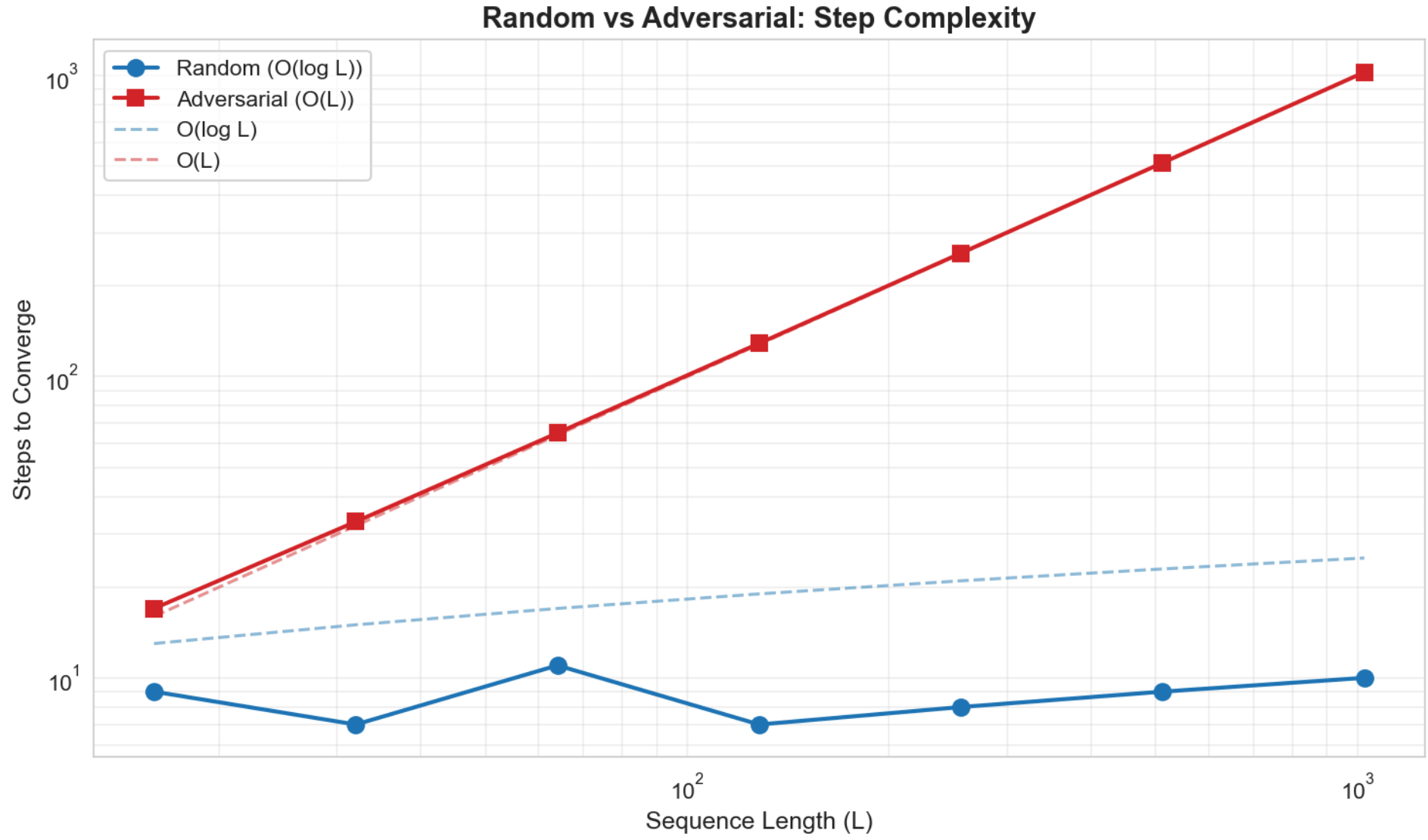

Figure 3: **Complexity analysis of convergence steps (Log-Log scale).** Horizontal axis is sequence length $L$ (log scale); vertical axis is steps to convergence (log scale). Blue line: Random input, convergence steps grow sub-$O(\log L)$. Red line: Adversarial input, convergence steps strictly linear in $L$, $O(L)$. Dashed lines are theoretical fits. This indicates SEAD spontaneously realizes the "Least Action" principle—dynamically adjusting computational resources based on problem difficulty.

The table below shows generalization results across different lengths. The metric is Exact Match (all output bits perfectly correct). For each length, we test both random and adversarial inputs ($1^L + 1$, worst case), and the model achieved 100% accuracy in all tests.

Table 3: Random vs Adversarial generalization results. All tests achieved 100% Exact Match accuracy.

| Length $L$ | Random (Steps) | Adversarial (Steps) |
|---|---|---|
| 16 | 9 | 17 |
| 64 | 11 | 65 |
| 256 | 8 | 257 |
| $1,024$ | 10 | $1,025$ |
| $10,000$ | 16 | $10,001$ |
| $100,000$ | 18 | $100,001$ |
| $1,000,000$ | 22 | $1,000,001$ |

It is worth noting that the model achieving such astounding generalization is extremely lean: only **99 trainable parameters**, trained for 5,000 steps on sequences of length $L = 16$, with a total training time of about 24 seconds. This scale is comparable to the Parity task (57 parameters), reaffirming SEAD's core argument—generalization capability comes not from parameter scale, but from correct inductive bias.

### 4.3 Probe III: Computational Universality (The Rule 110 Task)

Rule 110 is an Elementary Cellular Automaton introduced and numbered by Wolfram. In 2004, Matthew Cook rigorously proved that Rule 110 possesses Turing Completeness [18], meaning it can simulate any Turing machine and execute any computable process. The complexity of Rule 110 is manifested in its rich dynamic structures: Gliders, collisions, creation, and annihilation interactions.

The significance of Turing completeness lies in: if SEAD can accurately learn the transition rules of Rule 110, then SEAD itself inherits this computational power. In other words, SEAD is not just an "adder" or "parity checker", but a potential Universal Computational Engine—given the correct input encoding and local rules, it can theoretically execute any algorithm.

Consistent with the first two tasks, we use supervised learning to directly fit local rules. Specifically, we train the model on random initial states of $L = 32$, learning to predict the next state $h_{t+1}$ from the current state $h_t$. It is important to emphasize that the model learns under *supervised signals of known rules*, not discovering rules from chaotic data—the latter is the harder "rule discovery" problem, left for future work.

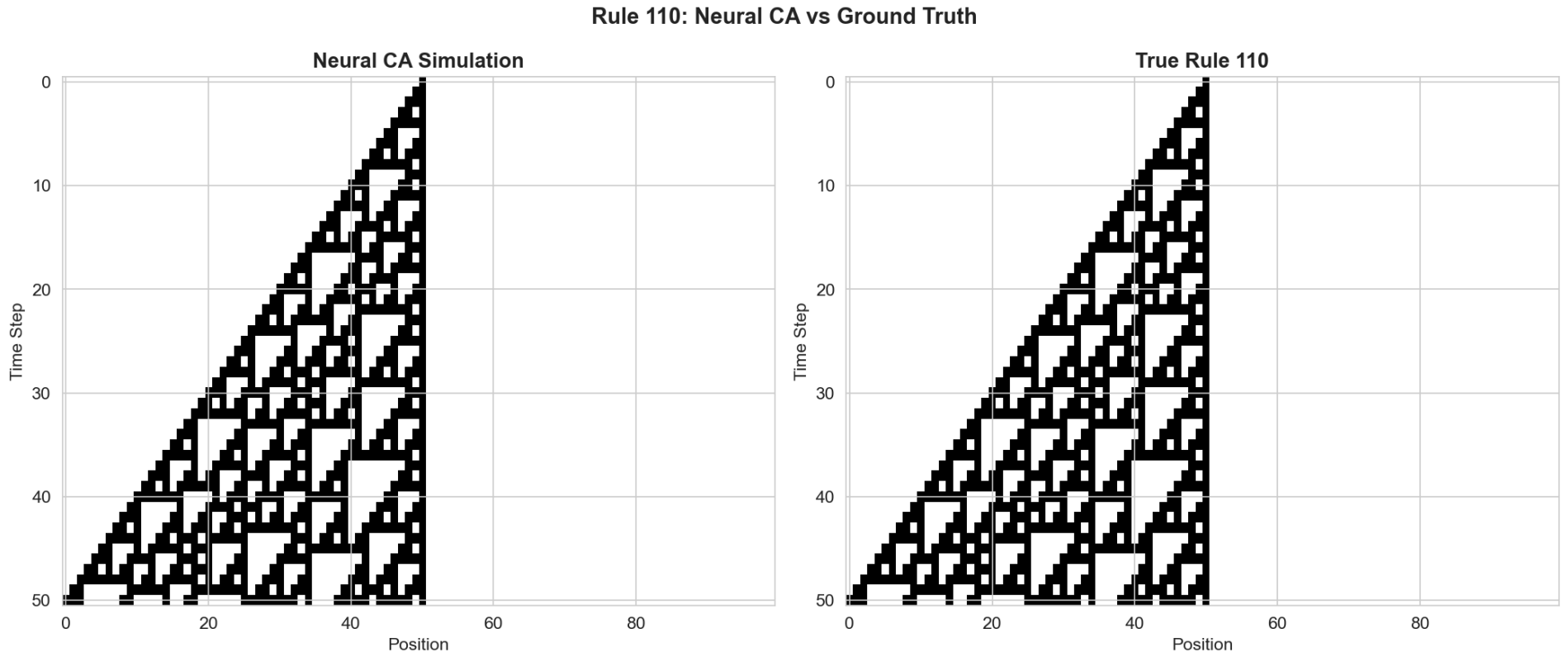

Figure 4: **Learning Rule 110: Neural Cellular Automata vs Ground Truth. Left**: Evolution graph generated by SEAD via supervised learning. **Right**: Evolution graph of true Rule 110. The two are visually identical, indicating SEAD has perfectly learned the transition rules and can losslessly simulate complex non-linear structures like Gliders and collisions.

Table 4: Rule 110 single-step prediction accuracy. Model trained on $L = 32$, tested on longer sequences.

| Test Length $L$ | Cell Accuracy | Sequence Accuracy |
|---|---|---|
| 32 | 100.0% | 100.0% |
| 64 | 100.0% | 100.0% |
| 128 | 100.0% | 100.0% |
| 256 | 100.0% | 100.0% |
| 512 | 100.0% | 100.0% |
| $1,000$ | 100.0% | 100.0% |

Experimental results show: The model achieves 100% single-step accuracy (both Cell Accuracy and Sequence Accuracy are 100%) across all test lengths. Since the inference mode uses deterministic argmax, 100% single-step accuracy guarantees zero cumulative error over multi-step evolution.

Since Rule 110 is Turing complete, and SEAD can precisely simulate Rule 110, this empirically indicates that the SEAD architecture itself possesses Turing completeness.

## 5 Discussion

### 5.1 The Space-like Fallacy of Transformers

In special relativity, the relationship between any two points in spacetime is strictly classified into two categories:

- **Time-like**: Two points can be connected by signals slower than light, allowing for causal relationship. Information can pass from A to B; B can be in A's "future". Physical laws evolve gradually along time-like paths via differential equations.
- **Space-like**: The distance between two points is too great and the time interval too short to be connected even by light. They are causally independent—A cannot affect B, nor B affect A.

The core of this classification is the *Light Cone*: from any point, only regions within its light cone can establish causal contact. Regions outside the light cone (space-like regions), while "physically existing", are causally inaccessible.

*Why is this relevant to Transformers?* Consider an addition problem: $111\cdots1 + 1$. The carry must propagate bit by bit from the least significant to the most significant—this is inherently a *Time-like* process, where each step depends on the result of the previous one. However, the Self-Attention mechanism allows arbitrary two positions (no matter how far apart) to interact directly, as if the entire sequence is presented "simultaneously" like an image. This compresses an infinitely extensible time-like process into a fixed-depth space-like map.

This is the "Space-like Fallacy" of Transformers: assuming logical reasoning can be parallelized, ignoring the temporal structure of causal chains.

This mindset directly leads to systematic violations of locality, symmetry, and stability:

1. **Violation of Postulate 1 (Locality)**: The Global Attention mechanism allows information from layer 1 to directly affect layer $L$. Physically, this is equivalent to assuming light speed $c \to \infty$. In this "infinite speed" pseudo-universe, the model loses the ability to distinguish "causal necessity" from "statistical correlation". This is why Transformers perform perfectly within $L_{\text{train}}$ (where correlation approximates causality within a short horizon) but collapse completely when $L_{\text{test}} \to \infty$ (where spurious correlations no longer hold over long distances).

2. **Violation of Postulate 2 (Symmetry)**: Transformers exhibit severe Spatiotemporal Heterogeneity. Spatially, Absolute Positional Encoding (APE) breaks translational invariance, making physical laws vary with coordinate $x$ (like assuming Newton's laws differ on different planets); temporally, non-shared weights between layers mean evolution rules vary with time $t$. In a universe where physical laws change arbitrarily with time and place, universal "generalization" clearly cannot exist.

3. **Violation of Postulate 3 (Stability)**: In a fully connected space-like graph, any local noise $\varepsilon$ can be instantly broadcast to the entire domain via the Attention matrix. The system lacks a "time-like" dissipative process to attenuate noise. In contrast, SEAD follows time-like evolution, where noise is confined within light cones during propagation and continuously corrected by attractors.

The success of the Transformer lies in utilizing GPU parallelism to simulate this "Space-like Universe". Its failure lies in the fact that the essence of logic is "Time-like"—consequences must follow causes and lie within their light cones. Forcing logic flat into spatial dimensions inevitably incurs the cost of fragile reasoning.

We acknowledge that the global attention mechanism of Transformers is highly effective for tasks requiring semantic non-locality (e.g., entity retrieval, coreference resolution in narrative text). SEAD is not proposed as a replacement for these semantic functions.

Instead, SEAD is designed as the specialized reasoning engine (System 2) that complements the Transformer's semantic engine (System 1). It restores the causal fidelity required for algorithmic generalization, which the standard Transformer fundamentally lacks.

### 5.2 The Empirical Validation of Physical Postulates

Our theoretical framework (Locality, Symmetry, Stability) is not built on air. In fact, every advancement in the Transformer community in recent years—whether theoretical insight or architectural innovation—can be interpreted as an unconscious approximation of these physical postulates. The community's success lies in partially restoring a postulate; their bottleneck lies in stopping there, failing to fully embrace physical constraints.

**Approximation to Postulate 1 (Locality)**: [10] significantly improved stability in arithmetic tasks by introducing Attention Bias or hard masks. [19] theoretically proved that Transformers can only achieve length generalization when each predicted token depends only on a fixed number of preceding tokens. These works essentially reconstruct locality manually within Transformers —the success of Duan et al. (though limited) is the best empirical proof of the value of locality. However, artificially introducing locality in a fully connected architecture is costly and limited.

**Approximation to Postulate 2 (Symmetry)**: Looped Transformers [20] and Universal Transformers [21] simulate recurrence by reusing layer weights, breaking the fixed depth limit. [7] and [8] designed ingenious Position Coupling and APE theories, attempting to make positional encoding translationally invariant. [9] reviewed extensive efforts in this direction. These works keenly realized that true reasoning requires temporal consistency (shared layer weights) and spatial equivariance (position independence). However, artificially introducing symmetry in an inherently heterogeneous architecture halts generalization at 2-3x.

**Approximation to Postulate 3 (Stability)**: Counteracting analog decay. Maintaining long-range information in continuous space faces the inevitable thermodynamic challenge—noise accumulation and signal decay. In Neural GPU, [22] introduced Hard Saturation and curriculum learning, forcing neuron states to tend towards discrete 0s and 1s during inference. This was an early and crude attempt at our "discrete attractor" concept—they realized that only by simulating the "hardness" of digital logic can computation be prevented from "softening" and failing over long sequences. Conversely, [23] recently proved the fundamental inability of State Space Models (SSMs), purely continuous systems, in length generalization, noting that "access to external discrete tools" is necessary to restore computational power. This inversely confirms Postulate 3: if a system lacks internal dissipative mechanisms to correct noise (like SEAD's attractors), it must rely on external sources of stability. True long-range reasoning must be built on a kind of "Digital Physics" capable of continuous self-correction.

**Chain-of-Thought: Revelation and Limitation**: Among all attempts, Chain-of-Thought (CoT) [11], [12] is the best performing method and deserves special discussion. CoT actually attempts to address all three postulates: for locality, it decomposes problems into local subtasks via step-by-step output; for symmetry, it implicitly introduces temporal recurrence by repeating similar reasoning steps; for stability, it explicitly externalizes intermediate states as tokens serving as external memory. The success of CoT proves the necessity of *Time-like Evolution*—the Transformer relies on an "external plugin" to simulate the recurrence its native architecture lacks.

However, CoT has fundamental flaws in all three directions. For locality, CoT tokens are still processed by Global Attention; early tokens can be accessed "at a distance" by later tokens, so locality is not truly restored. For symmetry, APE makes every CoT token positionally distinct; the model must learn that "adding at position 100" and "adding at position 200" are the same operation, destroying true translational invariance. For stability, CoT is autoregressive; once an early token errs, the error is read and amplified by subsequent tokens, with no intrinsic error-correction mechanism. More importantly, CoT is extremely inefficient: to perform a simple carry, the model might need to output thousands of tokens—simulating cheap CPU cycles with expensive I/O.

SEAD can be understood as Internalizing the CoT process into hidden layer dynamics. We do not need to spit out intermediate states as external tokens, but let them evolve naturally in the hidden state space. This internalization brings simultaneous satisfaction of efficiency, locality, symmetry, and stability.

Parallel to these architectural innovations, theoretical research is also delineating the boundaries of the old paradigm. [5] proved Self-Attention cannot model periodic languages and hierarchical structures—an inevitable consequence of *violating locality*: global attention cannot distinguish local causality from remote correlation. [6] proposed the "parallelism tradeoff": highly parallelizable architectures are necessarily limited to $\mathrm{TC}^0$ computational power—an inevitable consequence of *violating temporal symmetry*: without iterative evolution, computational depth is fixed. [17] proved the Transformer's loss landscape naturally biases towards low-sensitivity functions—another consequence of *violating locality*: high-sensitivity functions require precise coordination across the entire input, which global attention cannot provide as a structured constraint. Systematic studies by [2] on the Chomsky hierarchy, proofs of CoT necessity by [1] and [24], and characterization of CoT step lower bounds by [25] can all be unified under this framework. The community has realized the fundamental flaws of Transformers but lacks a unified theory to explain these seemingly unrelated limitations. Our three postulates provide exactly such a unified theory.

On another parallel track, Neural Cellular Automata (NCA) offer a computational paradigm naturally satisfying physical postulates. [26] proved the natural correspondence between CA and CNNs; [27] showed the self-organizing capability of Growing NCA; [28] reviewed NCA applications in biological morphogenesis and robotics. However, these works mostly focus on *image generation* (texture synthesis, pattern formation) or *reinforcement learning* (distributed control). [29] recently applied NCA to ARC-AGI reasoning tasks, demonstrating its potential.

Unlike these pioneering works, SEAD is among the few NCA works specifically targeting *Algorithmic Reasoning* and *Exact Logical Generalization*. We do not seek visual "approximate correctness", but mathematical "exactness"—100% accuracy from $L = 16$ to $L = 10^6$. This requirement for precision forces us to take physical postulates seriously, as any minor violation amplifies into catastrophic failure over long sequences.

Critics might argue that SEAD is structurally isomorphic to Neural Cellular Automata (NCA) or Neural GPUs. This is precisely the point.

SEAD fundamentally differs in its epistemology: We do not propose this architecture because it empirically works; we derive it as a mathematical necessity from physical postulates. The fact that our derived structure resembles previous heuristic attempts is not a lack of novelty, but a validation of Convergent Evolution—engineering intuition unknowingly stumbled upon the physical constraints that we have now rigorously formalized.

### 5.3 Future Directions: From Learning to Discovery

This study verifies the precise learning and generalization capabilities of SEAD under known physical rules. However, this is merely the starting point for neuro-symbolic physical systems.

When we examine the three tasks designed in this study, we find they all possess extremely simple logical rules understandable by humans. We proved SEAD can learn such simple rules under supervision and run losslessly. Although simple in form, these tasks (addition, CA rules) represent the Atomic Units of logical reasoning. Just as physicists studied the hydrogen spectrum before complex molecules, SEAD's lossless generalization on these foundational tasks proves it has mastered the axiomatic basis for building complex systems.

We naturally do not wish neural networks to stop at being an adder. In the future, we not only hope SEAD can learn more complex algorithmic logic, but more importantly, we hope it can learn real physical rules from real physical world data—for example, extracting explicit partial differential equations from learned convolution kernels and running them losslessly. This marks the fundamental shift of AI from "Fitter of Known Laws" to "Discoverer of Unknown Laws".

### 5.4 The Hardware Lottery

[30] proposed the "Hardware Lottery" hypothesis: the success of an algorithm often depends not on its theoretical superiority, but on whether it fits the current hardware environment. The Transformer is the ultimate winner of the GPU era because it converts all computation into Matrix Multiplication (GEMM), which GPUs excel at, masking its low sampling efficiency with extreme parallelism.

Regarding SEAD's hardware adaptability, we have two layers of thought:

1. **Compatible with Current Dividends**: In the training phase, SEAD decouples spatiotemporal evolution into independent local state transitions via Chaos Training. This means it remains a standard CNN, perfectly enjoying the parallel acceleration dividends of GPUs. We did not sacrifice training efficiency for causality. It should be noted that the training speed (seconds) in this paper relies on a premise: we possess dense supervision signals of local rules (an Oracle for single-step transitions). In practical applications with only end-to-end trajectory data, training might require Backpropagation Through Time (BPTT), increasing computational cost —but this cost is comparable to RNN or Neural ODE training, still significantly lower than the $O(N^2)$ attention overhead of Transformers. Meanwhile, we believe physical laws governing the complex world should be simple and clear, and learning simple laws does not require complex training processes. In the inference phase, the Transformer's $O(N^2)$ Global Attention consumes significant memory and power for long sequences, while SEAD's $O(N)$ local computation offers massive advantages in Edge AI / Robotics scenarios.

2. **Betting on Future Lotteries**: The space-like architecture of Transformers is tailored for SIMD hardware. However, the physical world is inherently local and asynchronous. With the development of Neuromorphic Chips and In-Memory Computing, we believe the future computational paradigm will return to locality. SEAD's "local computation, decentralized evolution" architecture naturally fits potential future "Causal Hardware". We are not just designing a model, but stockpiling algorithms for the next generation of computational substrates.

## 6 Conclusion

In this work, we have established a physics-grounded framework for understanding and achieving length generalization. We argued that the failure of current paradigms is not a matter of insufficient scale or clever engineering, but a systematic violation of fundamental physical postulates: Locality, Symmetry, and Stability. By enforcing these constraints, we derived—rather than designed—the SEAD architecture: a neural cellular automaton that inherits the computational properties of physical law. Our experiments demonstrate that SEAD achieves scale-invariant inference, generalizing perfectly from $L = 16$ to $L = 10^6$ with 100% accuracy—a feat impossible for models that rely on statistical correlation alone.

Finally, let us view all this from a broader historical perspective.

The Transformer empire is built on a Faustian bargain: trading the sanctity of time for the extreme efficiency of space. By collapsing the causal light cone into an undifferentiated global receptive field, it flattens events that must occur sequentially in reality into a simultaneously presented panorama. This "God's eye view" parallelism released immense energy in language modeling—a task of strong correlation and weak causality—launching the golden age of Large Language Models.

However, as the next frontier of AI shifts from "understanding language" to "logical reasoning" and "scientific discovery", the dividends of yesterday become the debts of today. Reasoning is not a panorama, but a path. Science is not correlation, but causation. The nature of these tasks is *Time-like*—they respect the arrow of time and obey the constraints of light cones. The Space-like architecture of the Transformer runs precisely counter to this.

"Attention Is All You Need"—this slogan may be more profound than its inventors realized. It is not just a manifesto, but a prophecy. For in a universe universally dominated by Global Attention, *nothing truly happens "after" another*. Everything is seen in an instant, meaning nothing can be truly "derived". Attention is indeed all you need—until you need reasoning.

Even Chain-of-Thought serves only as a ritual to simulate causality within an acausal machine. It attempts to reconstruct the arrow of time, using explicit token sequences to feign a reasoning process. Yet, a specter is haunting the reasoning tokens—the specter of Global Attention. As long as the architecture allows 'looking ahead,' true physical generalization remains a mirage. The Transformer succeeded by ignoring the constraints of physics; it will reach its limit because physics cannot be ignored forever. *It rose by shattering time; it shall fall for want of it.*

We need no more Global Attention—because the physical world itself is *Local*.

We need no more elaborate Positional Encodings—because physical laws themselves are *Coordinate-Independent*.

We need no Chain-of-Thought to externalize time—because intelligence itself is a *Function of Time*.

SEAD is not a competitor to the Transformer; it is the *successor* to the Transformer in the domain of reasoning. It reclaims the physical puzzle pieces discarded by Global Attention: Locality gives it causal structure, Symmetry gives it generalization power, and Stability gives it long-range memory. It brings intelligence back from the starry sky of pure mathematics to the solid ground of physical reality.